\documentclass[conference]{IEEEtran}
\IEEEoverridecommandlockouts
\usepackage{cite}
\usepackage{amsmath,amssymb,amsfonts}
\usepackage{algorithmic}
\usepackage{graphicx}
\usepackage{textcomp}
\usepackage{xcolor}
\usepackage{multirow}
\usepackage{float}
\usepackage{tikz}
\usepackage{booktabs}
\usetikzlibrary{chains}

\def\BibTeX{{\rm B\kern-.05em{\sc i\kern-.025em b}\kern-.08em
    T\kern-.1667em\lower.7ex\hbox{E}\kern-.125emX}}
\begin{document}

\title{A two-way translation system of Chinese sign language based on computer vision}

\author{\IEEEauthorblockN{1\textsuperscript{st} Shengzhuo Wei}
\IEEEauthorblockA{\textit{Harbin Institute of Technology} \\
Harbin 150001, China \\
Cherrling2021@outlook.com}
\and
\IEEEauthorblockN{2\textsuperscript{nd} Yan Lan}
\IEEEauthorblockA{\textit{Communication University of China } \\
BeiJing 100024, China  \\
lanyan761091450@gmail.com}
}

\maketitle

\begin{abstract}
As the main means of communication for deaf people, sign language has a special grammatical order, so it is meaningful and valuable to develop a real-time translation system for sign language. In the research process, we added a TSM module to the lightweight neural network model for the large Chinese continuous sign language dataset . It effectively improves the network performance with high accuracy and fast recognition speed. At the same time, we improve the Bert-Base-Chinese model to divide Chinese sentences into words and mapping the natural word order to the statute sign language order, and finally use the corresponding word videos in the isolated sign language dataset to generate the sentence video, so as to achieve the function of text-to-sign language translation. In the last of our research we built a system with sign language recognition and translation functions, and conducted performance tests on the complete dataset. The sign language video recognition accuracy reached about 99.3\% with a time of about 0.05 seconds, and the sign language generation video time was about 1.3 seconds. The sign language system has good performance performance and is feasible.
\end{abstract}

\begin{IEEEkeywords}
Chinese sign language,sign language recognition, sign language generation, language model, Transformer
\end{IEEEkeywords}

\section{Introduction}
\subsection{Background}
While able-bodied people can use verbal language to communicate easily, people with hearing impairment (deaf or aphasic people, etc.) need to communicate their thoughts through sign language. There are about 20.57 million deaf people in China, accounting for 1.67\% of the total Chinese population, including about 800,000 children under the age of 7. They cannot communicate through language as normal people do, but communicate through sign language.

Since most of the able-bodied people have not learned sign language, there are obstacles to promote sign language to make it applicable for communication in normal society. Sign language recognition and interpretation technology facilitates communication between hearing-impaired and able-bodied people. Sign language research should not only enable hearing people to read sign language, but also enable hearing people to understand what able-bodied people are saying. 

Sign language recognition and interpretation are the former, and sign language generation research is the latter. This interaction process is particularly important for people with hearing impairment. 

Therefore, the study of sign language recognition and interpretation and sign language generation has important theoretical and applied values as well as social significance. Sign language recognition technology and sign language generation technology can help daily communication, sign language interpretation and sign language education activities between deaf and able-bodied people, as well as improve the social skills and quality of life of deaf people, promote mutual understanding and communication between deaf and able-bodied people, and have practicality and applicability in the deaf community.

\subsection{Related work}
\subsubsection{Current status of research on sign language recognition}
The sign language recognition method based on wearable devices, i.e., using data gloves to directly obtain the hand shape, angle and relative position of fingers and other precise data of the granting person, so as to obtain the main characteristics of sign language and use recognition algorithms for recognition. This method does not require pre-processing of various information, and the acquired data is accurate and free from environmental interference, with the disadvantage of high cost and complexity of use.

Computer vision-based sign language recognition method, i.e., the sign language gesture image or dynamic change information is obtained by camera or radar and input to the algorithm for sign language recognition. Compared with sign language recognition based on wearable devices, sign language personnel do not need wearable devices and the promotion is more advantageous. The disadvantage is that the exclusion of fuzzy frames, the pre-processing of data to exclude interfering information and the accuracy of information is not high. 

Table I shows the current status of research on sign language recognition in China and abroad.
\begin{table}[H]
\caption{Current status of sign language recognition research}
    \centering
    \begin{tabular}{cccc}
    \toprule
        Research Team&Data collection method&Identify the content&Recognition accuracy\\ 
        \midrule
        Kadous&PowerGlove data glove&95 Australian Sign Language Words&80\%  \\  
        Kakoty&Sensor Stacking Gloves&Simple one-handed sign language&96.7\% \\
        Jiang Li&CAS-Glove Data Glove&Common Sign Language Words&97.2\% \\
        Wang Jincheng&Data gloves and smartphones&Simple Sign Language&95\% \\
        Foreign Researcher Team&Kinect depth information&Common sign language words or simple sentences&95\%\\
        Zhang Liangguo&New TMHMM Model&439 common words in sign language&92.5\%\\
        Liu Yaling&VGG-Net convolutional neural network&22 letters of the alphabet&97\%\\
        \bottomrule
    \end{tabular}
\end{table}

\subsubsection{Current status of research on sign language generation}
With the continuous update of deep neural network theory, the update of intelligent devices and the development of computer technology, inspired by various generative models, researchers around the world have provided several methods or systems for solving the problem of sign language generation.

Table II shows the current status of research on sign language generation in China and abroad.

\begin{table}[H]
\caption{Current status of research on sign language generation}
    \centering
    \begin{tabular}{ccc}
    \toprule
        Research Team&Generating method&Applicable direction\\ 
        \midrule
Greek scholar Karpouzis	&Standard virtual character animation technology	&Virtual animation of its signature\\
Glauert	&Compositing to generate body animations&	Speech-driven for e-government\\
Saunders &	Confrontational multi-channel SLP approach	&Generate facial features and mouth patterns\\
Ben&Improved Transformer Generator	&Pose video generation using conditional GAN networks\\
Stephanie	&Neural machine translation	&Generating gesture videos from sign language word sequences\\
Jan Zelinka	&Pose Sequence Generation Framework	&Train a sign language generation network using this skeleton data\\
\bottomrule
    \end{tabular}
\end{table}

In summary, domestic and foreign research scholars have proposed many methods for sign language generation, however most of the generation methods are useful for generating sequences containing temporal information, on the other hand Chinese sign language is a sign language expression different from other languages, and the generation methods proposed by research scholars for Japanese, English or Greek cannot be directly used for the generation of Chinese sign language, and the models are not universal. To address the above problems this paper proposes a deep learning-based sign language recognition and sign language generation method and provides a system that can be used for two-way communication between deaf and normal people, which can be extended for other scenarios.

\subsection{Research Content}
This paper describes the importance of sign language in the communication between deaf and able-bodied people, and presents the research of Chinese sign language recognition algorithm and sign language generation algorithm based on deep learning and the implementation of the system.

Deep learning is an artificial neural network technique that can learn and classify a large amount of data. In sign language recognition technology, deep learning technology can learn and classify a large amount of sign language images and video data to achieve recognition and translation of sign language. The research of Chinese sign language recognition technology and sign language generation technology and system based on deep learning can provide more accurate and efficient sign language recognition and translation services for communication between deaf and able-bodied people.

In this paper, we combine different behavior recognition depth models, study the advantages and shortcomings of currently existing sign language recognition algorithms, delve into the latest research results of deep learning at home and abroad, consider the structural characteristics and structural advantages of existing convolutional neural network models, and explore the applicability, stability, and reliability of different structural convolutional neural networks in the field of sign language recognition.

In the part of sign language recognition algorithm research, this paper firstly conducts video preprocessing, video feature extraction on the Chinese sign language continuous utterance SLR dataset constructed by the University of Science and Technology of China, and then adds the TSM module with better processing capability of temporally strongly correlated video to ResNet-50 and MobileNet models for training sign language video and analyzing experimental data; meanwhile, conducts Action-net, another type of behavior recognition model, is experimented to recognize sign language videos and analyze the experimental data; through experimental comparison, the performance and effect of the models are verified, and a sign language recognition algorithm suitable for CSL continuous utterance dataset is selected to realize the recognition and translation of sign language continuous utterances. 

In the part of sign language generation, the research aims to transform the spoken Chinese text into the sequence of sign language language through jieba splitting and then processed by Bert model, corresponding to the corresponding sign language vocabulary video, to complete the translation from text to video, and to verify the performance and effect of the model.

\section{Method}
\subsection{System Structure Design}
This project is dedicated to designing a two-way sign language translation system, which is mainly divided into sign language recognition part and sign language generation part, and the main flow chart of the two parts is as the following Figure1.
\begin{figure}[H]
    \centering
    \includegraphics[width=6in]{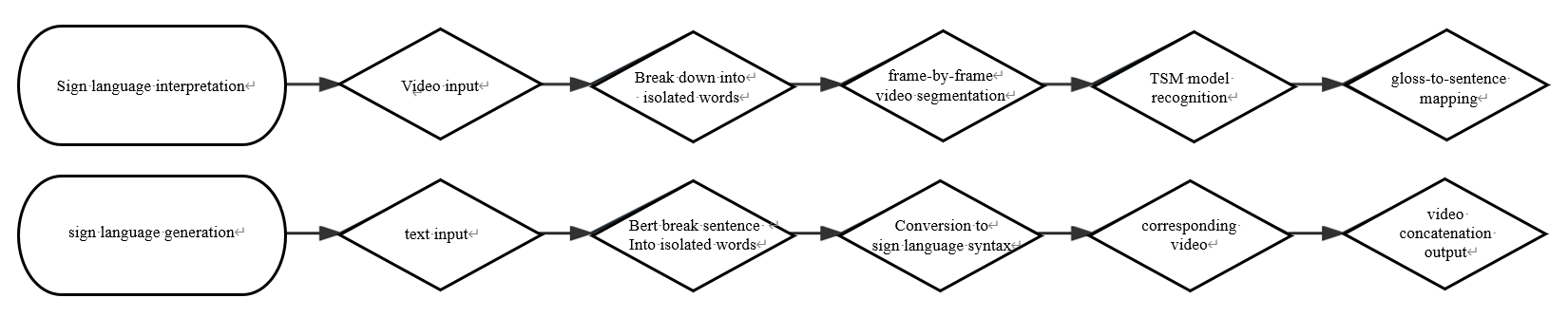}
    \caption{Flow chart of sign language two-way translation system}
    \label{fig:my_label}
\end{figure}

In this system, since sign language recognition translation is a temporal strong correlation process, TSM in behavior recognition model with Action-net is used for correlation model training.

In this system, the sign language generation part relies heavily on the natural corpus segmentation and processing, so the Bert pre-training model, which is more efficient and accurate, is chosen as the basis for development.

\subsection{Sign language dataset}
The correspondence rules between sign language grammar and natural language grammar are important indicators for the selection of sign language datasets. According to the study of sign language linguistics, usually the sign languages of some countries can be divided into natural sign language and statute sign language (or gestural sign language). Taking Chinese sign language as an example, Chinese sign language can be divided into natural sign language and gestural Chinese. Natural sign language is mainly used by the hearing impaired and has a set of systematic grammatical rules, while gestural Chinese is an artificial language that operates directly on the basis of spoken grammar with gestures and has a one-to-one correspondence with Chinese characters, and is therefore also called written sign language.

How to map natural sign language and statute sign language is one of the challenges of sign language translation research. Most of the existing research on sign language translation is based on continuous sign language recognition, combined with language models to obtain natural language translations that conform to spoken descriptions. In the future, we can consider constructing large text pair datasets, i.e., the natural sign language annotation set and the corresponding statute sign language annotation set, and pre-training the language model on the text pair dataset first, and then migrating it to the language model of sign language translation.

In the continuous sign language utterance dataset, a portion of the dataset has annotations against the text in normal spoken language order, which can be used for sign language generation translation functions, mainly including Boston-104, RWTH-PHOENIX-WEATHER-2014-T, KETI, GSL, MEDIAPI-SKEL corpus, and CSL-Daily datasets. The CSL-Daily dataset can be used for continuous sign language recognition and translation tasks, and it provides spoken language translation and lexical level annotation. Compared with USTC-CCSL, CSL-Daily focuses more on daily life scenarios, including multiple topics such as family life, healthcare and school life. the training, validation and test sets of CSL-Daily contain 18,401, 1,077 and 1,176 video samples, respectively. Figure2 shows the main struct of the dataset and the Table III shows the difference between these datasets.

\begin{figure}[H]
    \centering
    \includegraphics[width=0.5\linewidth]{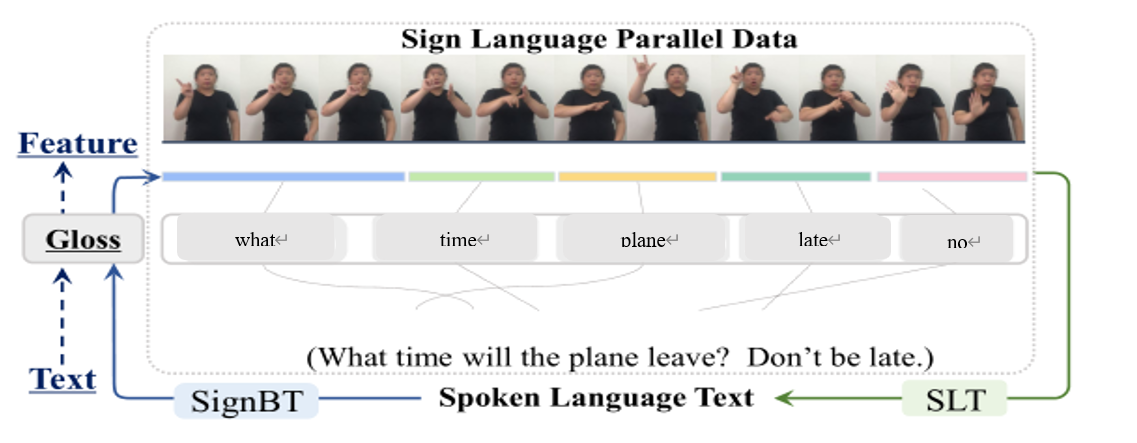}
    \caption{ Relevant composition of the sign language dataset}
    \label{fig:my_label}
\end{figure}

\begin{table}[H]
\caption{Overview of Chinese sign language dataset}
    \centering
\begin{tabular}{cccccccc}
\toprule
Data set name & Year & China& Type&Category &Sample Examples&Recorders Numbers& Data Characteristics \\
\midrule
DEVIDIGN-G& \multirow{3}{*}{2014} & \multirow{3}{*}{China}&\multirow{3}{*}{Isolated words}  &36 & 432& \multirow{3}{*}{8} & \multirow{3}{*}{RGB} \\
DEVIDIGN-D&  & &  &500 & 6000 & &  \\
DEVIDIGN-L &  & &  &2000 & 24000 &  & \\

\multirow{3}{*}{USTC-CCSL}&\multirow{3}{*}{2015}  &\multirow{3}{*}{China} & Isolated words&500 & 125000 & \multirow{3}{*}{50}& RGB, depth, skeleton \\
 & & & Continuous &100 &25000  & & RGB \\
 & & &Statements&100&25000 &  & RGB \\
\bottomrule

\end{tabular}
\end{table}

\subsection{TSM Model}
Time Shift Module (TSM), which provides a new approach for effective temporal modeling in video understanding, is an enhanced derivative model for temporal information learning based on the TSN model. In the TSN model, temporal information is fused by taking N images from the video relatively equally and randomly, and then averaging their classification results to achieve a certain degree of temporal information modeling. The TSM model absorbs the advantages of the TSN model, and at the same time, after the selection of each image, uses the shift of the time dimension to enable a single image to contain the information features of multiple neighboring images, thus greatly improving the efficiency of temporal information recognition, and at the same time, because only part of the image information is shifted for information aggregation, feature fusion between different frames can theoretically be achieved on the basis of zero additional computational overhead Joint modeling, which can be computed freely on the basis of two-dimensional convolution, but has a strong time-domain modeling capability. Figure 3 explains the main principles of the TSM model for frame shifting.
\begin{figure}[H]
    \centering
    \includegraphics[width=0.5\linewidth]{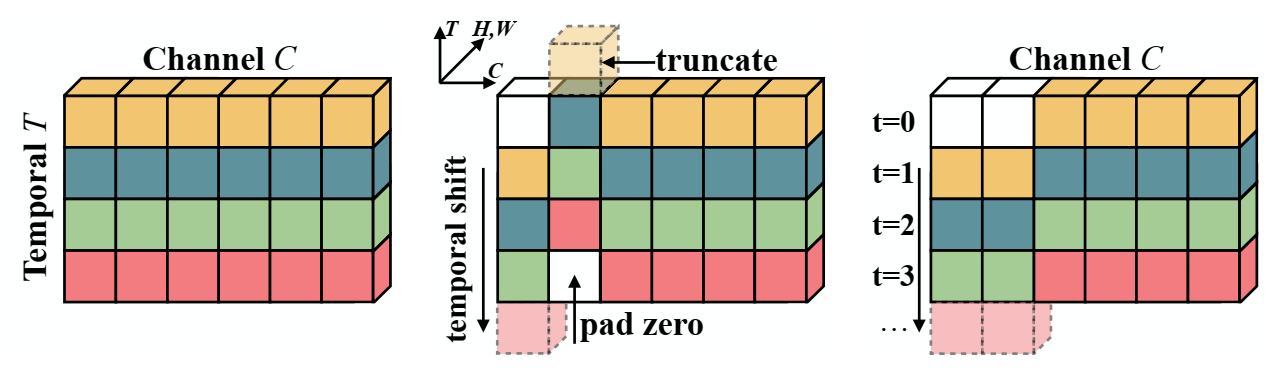}
    \caption{TSM model core principle - time series displacement}
    \label{fig:enter-label}
\end{figure}
TSM performs efficient temporal modeling by shifting feature mapping along the temporal dimension.TSM efficiently supports offline and online video recognition. Two-way TSM mixes past and future frames with current frames for high-throughput offline video recognition. One-way TSM blends only past frames with current frames and is suitable for low-latency online video recognition.

As shown in the figure, one-way TSM online video recognition requires features from future frames to replace the features in the current frame. One-way TSM online recognition can be achieved by simply transferring features from the previous frame to the current frame. The one-way TSM inference diagram for online video recognition is shown in Figure 4.
\begin{figure}[H]
    \centering
    \includegraphics[width=0.5\linewidth]{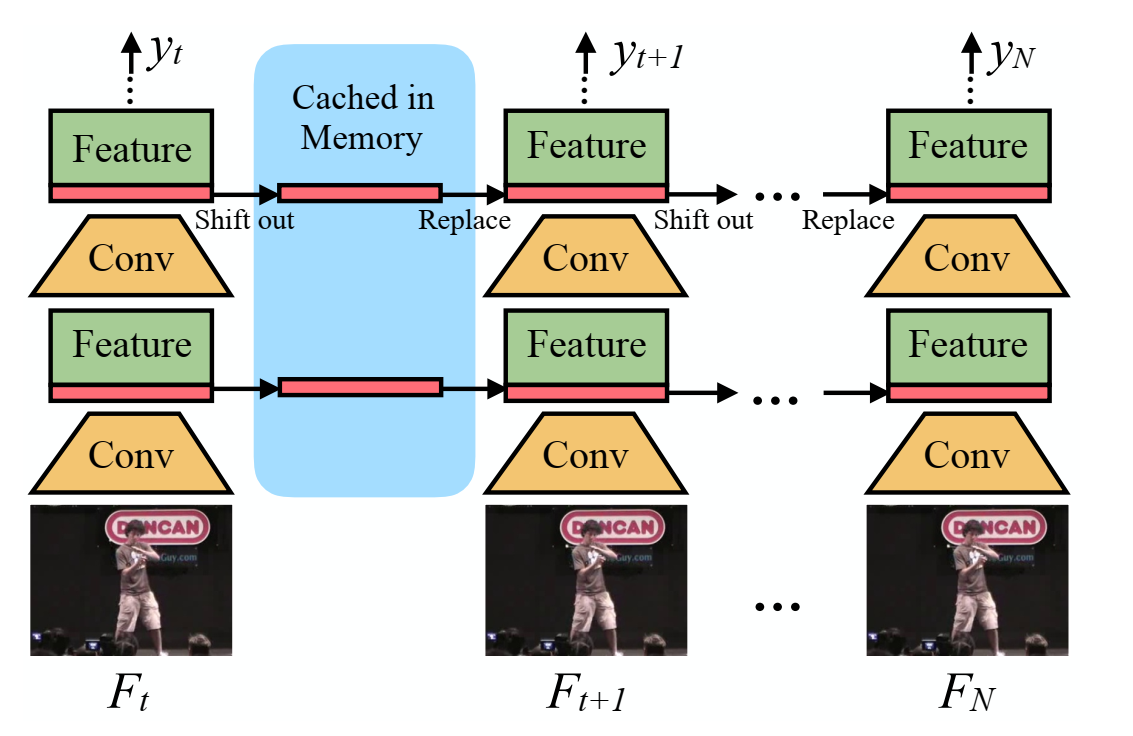}
    \caption{Time-shift operation performed by TSM on the actual video}
    \label{fig:enter-label}
\end{figure}

During the inference process, for each frame, we save the first 1/8 feature mapping of each residual block and cache it in memory. For the next frame, we replace the first 1/8 of the current feature mapping with the cached feature mapping. We use a combination of 7/8 of the current feature mapping and 1/8 of the old feature mapping to generate the next layer, and repeat. Using one-way TSM for online video recognition has several unique advantages: low latency inference.
TSM is an enhanced derivative model for temporal information learning based on the TSN model.
In the TSN model, the temporal information is fused by taking N images from the video relatively equally and randomly, and then averaging their classification results to achieve a certain degree of temporal information modeling.

The TSM model absorbs the advantages of the TSN model, and at the same time, after the selection for each image, uses the shift of the temporal dimension to enable a single image to contain the information features of multiple neighboring images, thus greatly improving the efficiency of temporal information recognition, and at the same time, because only part of the image information is shifted for information aggregation, feature fusion between different frames can be achieved in theory on the basis of zero additional computational overhead Joint modeling.

The TSM model part of this project was trained on a cloud computing platform using 5 vCPU Intel(R) Xeon(R) Silver 4210R CPU @ 2.40GHz with RTX 3090 (24GB) * 1. The training was conducted in about 30 hours for SLR-100 sentence dataset and 46 hours for SLR-500 word dataset.
    
\subsection{ Action-net model}
Traditional 2D CNNs are computationally inexpensive but cannot capture temporal relationships. In contrast, 3D CNN can capture the temporal relationships but is computationally expensive. In the design of Action-Net, it consists of three attention mechanism modules, including the Spatial-Temporal Excitation module, the Channel Excitation module and the Motion Excitation module, so that it can be more effective for the temporally strongly correlated The video framing is better than the video framing.
The format of the image is [N,T,C,H,W], where N denotes batch size, T denotes number of segments, C denotes number of channels, H denotes height height, W denotes width, and r is the channel loss rate channel reduce radio. Figure 5 illustrates the main structure of the Acrion-net module.
\begin{figure}[H]
    \centering
    \includegraphics[width=0.75\linewidth]{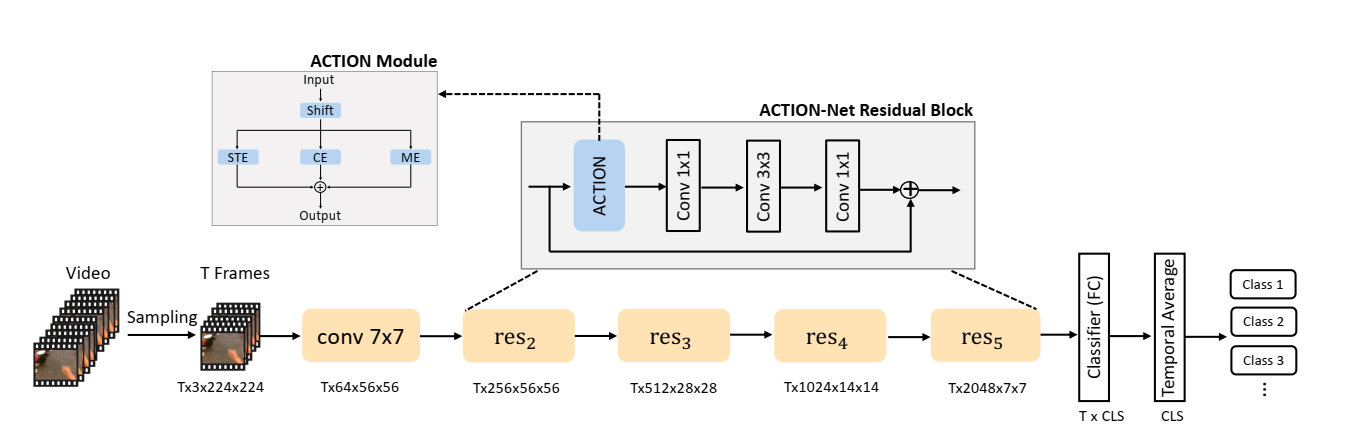}
    \caption{The main structure of the Acrion-net module}
    \label{fig:enter-label}
\end{figure}

ACTION module: The ACTION module is made up of the three attention modules mentioned above in parallel. This module is plug-and-play like the previous work TSM. The base model uses the same ResNet-50 as in the previous work.
The sign language recognition and translation process designed in this project belongs to the temporal strong related project, meanwhile, since TSM and Action-net perform well in behavior recognition and there are more related materials, this project uses TSM and Action-net models for sign language translation related training
Traditional 2D CNNs are computationally inexpensive but cannot capture temporal relationships. While 3D CNN can capture the temporal relationship, but it is computationally more expensive.
In the design of Actionnet, the Spatial-Temporal Excitation module, the Channel Excitation module, and the Motion Excitation module are proposed. Finally, they are added together to implement the Action module. Thus, it has a good effect on the video frame splitting with strong timing correlation.
The Action model natively supports jester, sth-sth v1, v2 and many other mainstream datasets. Considering the file structure and other issues, we finally chose to process the existing SLR dataset into jester standard format.
The processing starts with frame-slicing the video from the existing dataset using tools such as ffmpeg. After the video is frame-separated, a python-based script is used to count the number of frames, RGB and optical flow information of the frame-separated file. Since the Action native code uses pkl files to save the tag information, the Pickle library is called to save the corresponding path information and tag information to the corresponding tag pkl files after distinguishing the training set and the test set.

In this project, the TSM model was partially trained on a cloud computing platform using 15 vCPU AMD EPYC 7543 32-Core Processor with RTX A5000 (24GB) * 1, where the SLR-100 sentence dataset was trained for about 20 hours.

\subsection{Bert model (bert-base-Chinese)}
Bert is an unsupervised pre-trained language model for natural language processing tasks. The goal of the Bert model is to use a large-scale unlabeled corpus to train and obtain a textual semantic information-rich The goal of the Bert model is to train a large-scale unlabeled corpus, obtain a semantic representation of the text containing rich semantic information, and then fine-tune the semantic representation of the text for a specific NLP task and finally apply it to that NLP task.

Work on sign language generation relies heavily on the analysis and segmentation of natural language, and Bert has a good efficiency and performance in processing natural corpus. Therefore, in this project, a pre-trained bert-base-Chinese model is used as the basis, and further training and tuning is performed on top of it to realize the process of segmenting and adapting natural language into sign language sequences.
  
In the sign language video-to-annotation-to-text based sign language translation paradigm, the sign language translation process is divided into two phases: the first phase treats sign language recognition as an intermediate tokenization component that extracts sign language annotations from the video; the second phase is a language translation task that maps the sign language annotations to spoken text.

In the processing of the CSL-Daily dataset, its video-map is first extracted to select the corresponding information of the useful natural and sign language sequences. Then it is converted into a format suitable for training the bert model.
In training, the model first reads the original sentences of each data item, and then compares them with the validated sentences after the word separation process.

The experiments in this chapter are built on the datasets corresponding to large sign language texts, i.e., the natural sign language annotation set and the corresponding statute sign language annotation set, and the language model is first pre-trained on the text pair dataset and then migrated to the language model for sign language translation. The input spoken text is taken through the Bert-Base-Chinese model, and the corresponding sign language video is automatically output by the system, as shown in Figure 6:

\begin{figure}[H]
    \centering
    \includegraphics[width=0.5\linewidth]{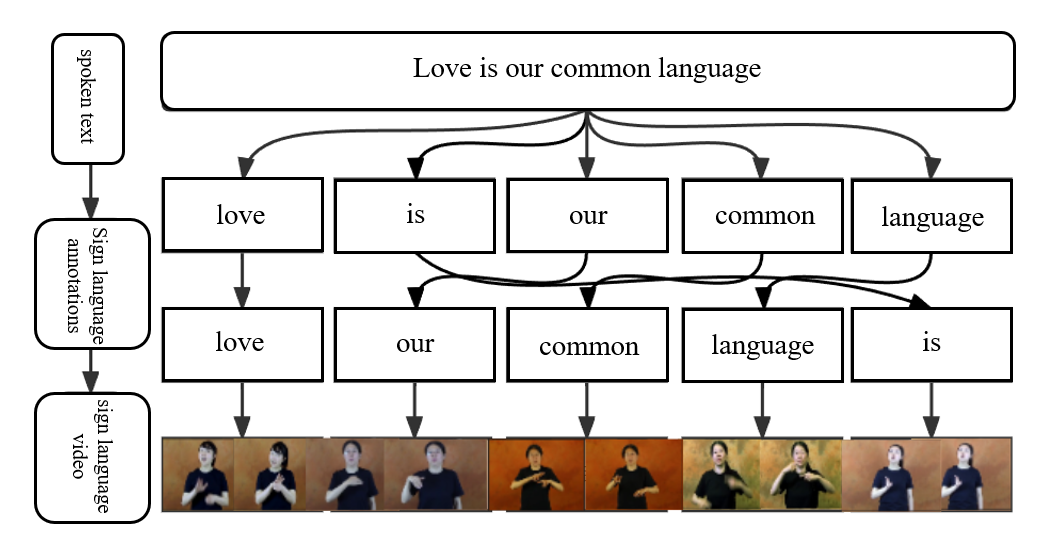}
    \caption{Basic principles of sign language generation}
    \label{fig:enter-label}
\end{figure}

This experimental model is partially trained on a cloud computing platform using a 5 vCPU Intel(R) Xeon(R) Silver 4210R CPU @ 2.40GHz with RTX 3090 (24GB) * 1 for training.

The Bert model can already be used for natural language word separation, and is trained on top of the existing Bert-base-Chinese pre-training model by importing CSL-Daily related data sets. The purpose of this training is to output the words generated by the model in sign language sequence for subsequent generation of sign language videos, and to complete the spoken text-syntax-sign language annotation. Figure 7 illustrates the main steps of sign language segmentation using the Bert model.

\begin{figure}[H]
    \centering
    \includegraphics[width=0.75\linewidth]{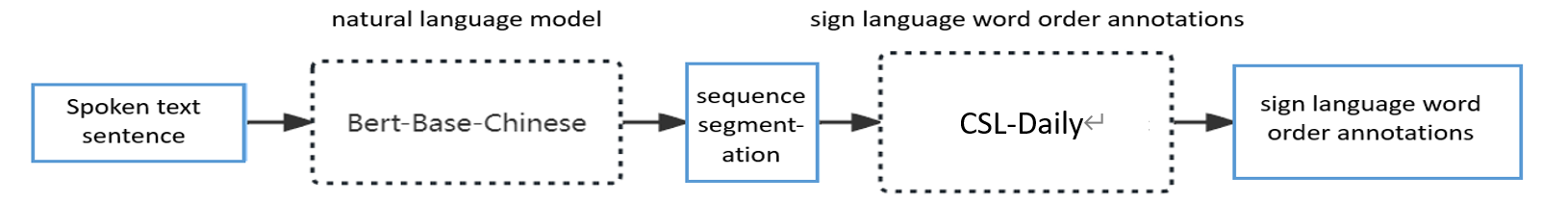}
    \caption{ Flow chart of sign language generation}
    \label{fig:enter-label}
\end{figure}
\section{Results}
\subsection{TSM experimental results}
In the training for 100 consecutive sentences, the following results were obtained after 117 iterations based on resnet50. Figure 8 shows the training results of TSM based on resnet50 for 100 sentences.

Testing Results: Prec@1 99.300 Prec@5 99.960 Loss 0.02676
\begin{figure}[H]
    \centering
    \includegraphics[width=0.5\linewidth]{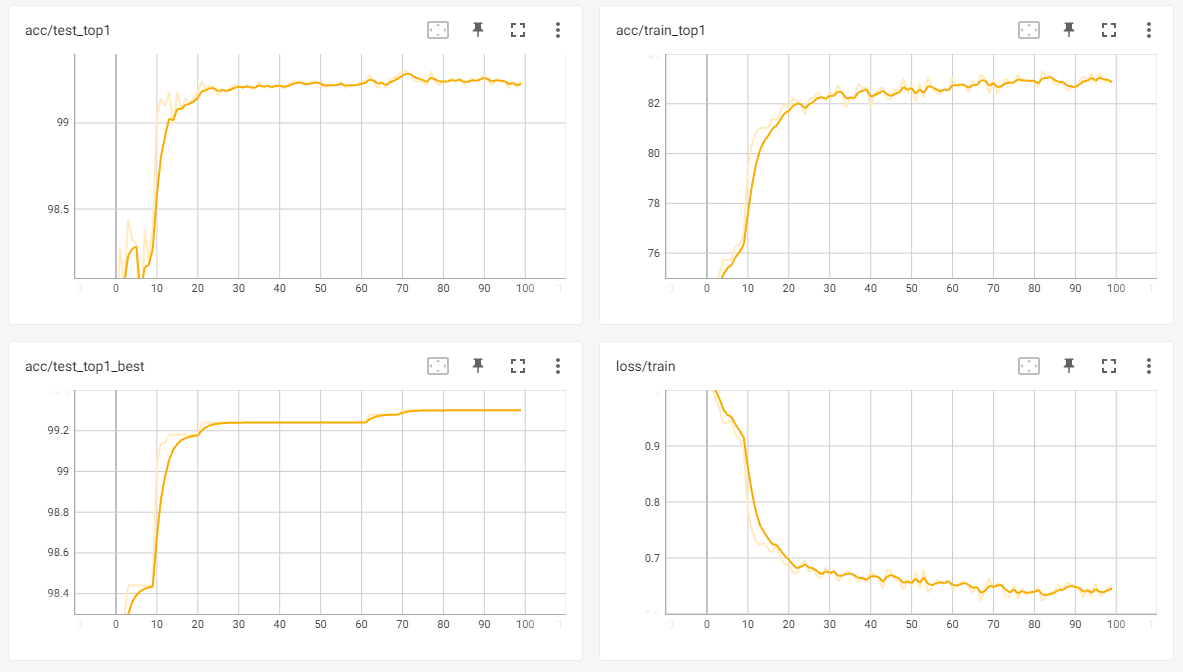}
    \caption{Training results of TSM based on resnet50 for 100 sentences}
    \label{fig:enter-label}
\end{figure}
In the training of 500 isolated sign language words, the following results were obtained after 30 iterations based on resnet50. Figure 9 shows the training results of TSM based on resnet50 for 500 isolated words

Testing Results: Prec@1 96.840 Prec@5 96.840 Loss 0.11426
\begin{figure}[H]
    \centering
    \includegraphics[width=0.5\linewidth]{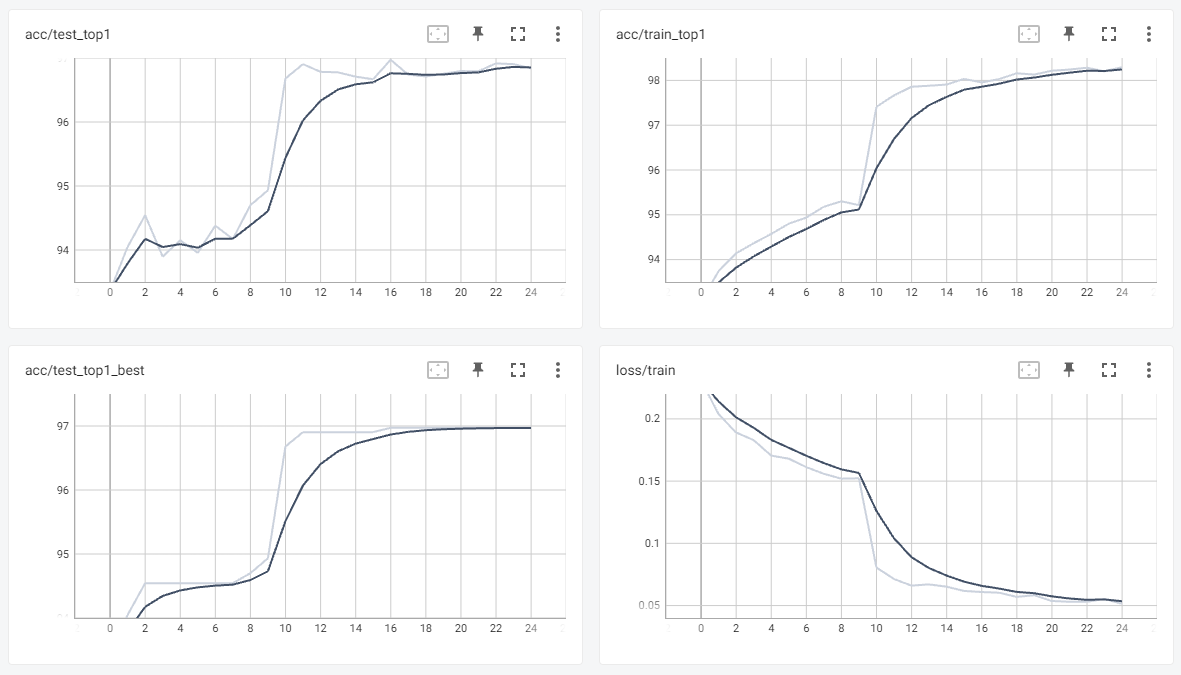}
    \caption{Training results of TSM based on resnet50 for 500 isolated words}
    \label{fig:enter-label}
\end{figure}
In the training for 100 consecutive sentences, the following results were obtained after 117 iterations based on mobilenet. Figure 9 shows the training results of mobilenet-based TSM for 500 isolated words.

Testing Results: Prec@1 95.240 Prec@5 95.240 Loss 0.17194
\begin{figure}[H]
    \centering
    \includegraphics[width=0.5\linewidth]{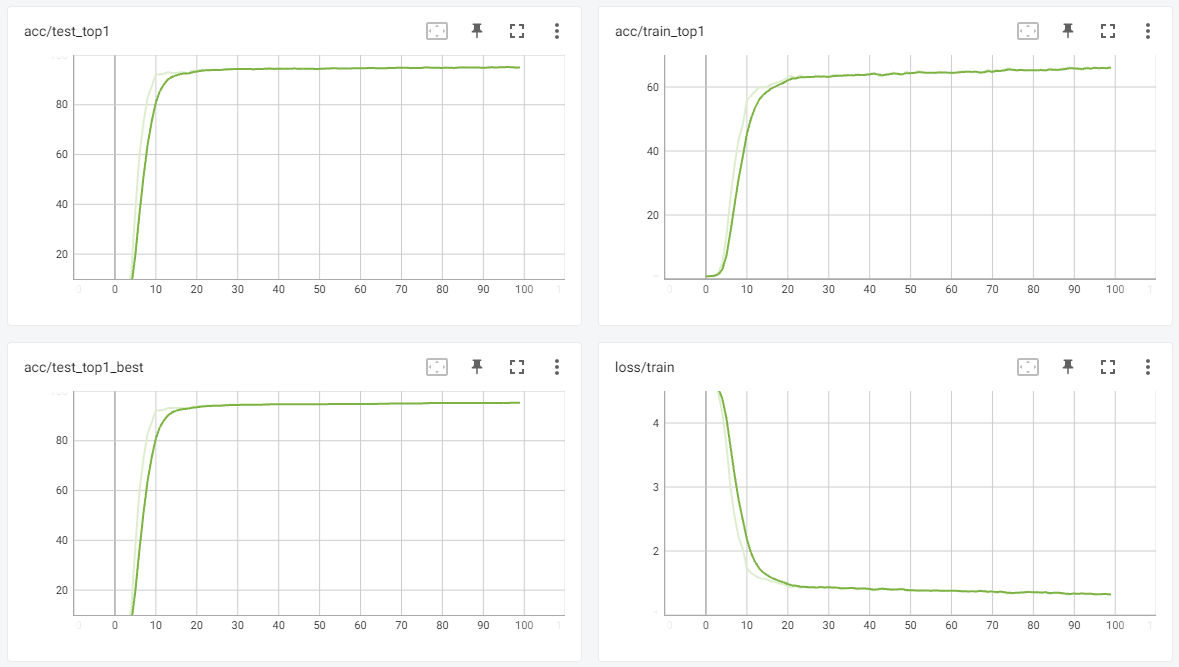}
    \caption{Training results of mobilenet-based TSM for 500 isolated words}
    \label{fig:enter-label}
\end{figure}
\subsection{Action-net experiment results}
In the training for 100 sentences, the following results were obtained after 48 iterations.

Testing Results: Prec@1 34.580 Prec@5 62.050 Loss 2.6734

\subsection{Comparison of TSM and Action-net experiments}
After 48 iterations of training in TSM and Actionnet respectively, the results are as the follow Table IV.

\begin{table}[H]
\caption{Comparison of TSM and Action-net training results}
    \centering
\begin{tabular}{cccc}
\toprule
&Prec@1&	Prec@5	&Loss \\
\midrule
TSM-resnet50&	97.720 &	99.740&	0.07499 \\
Actionnet	&34.580	&62.050&	2.6734 \\
\bottomrule
\end{tabular}
\end{table}

Considering the model accuracy and model convergence speed, TSM is used as the implementation model in this project.

Also in the TSM training based on different underlying models, after 100 iterations each, the results are as the follow Table V:

\begin{table}[H]
\caption{Comparison of TSM training results based on resnet50 and mobilenet}
    \centering
\begin{tabular}{cccc}
\toprule
&Prec@1&	Prec@5	&Loss \\
\midrule
TSM-resnet50&	97.720 &	99.740&	0.07499 \\
TSM-mobilenet	&95.240&	95.240&	0.17194 \\
\bottomrule
\end{tabular}
\end{table}
Considering the model accuracy and model convergence speed, and due to the poor performance of mobilenet in the post-test, the TSM model with resnet50 as the underlying layer was finally adopted for the target implementation in this project.

\subsection{Sign Language Interpretation Test Results}
The test for the sign language translation module of this project is mainly divided into two parts: continuous sentence recognition and isolated word recognition, and the main test contents are as follows.

The video of continuous sign language sentences was obtained from the friendly assistance of students from the sign language club of HIT, and the video of isolated word sign language was obtained from the public sign language information website and related data sets.

Tested on the trained model, both recognition results are better, and the isolated word sign language recognition model will be mainly used in the follow-up.

\subsection{Bert experimental results}
After iterating a total of about 23k steps over a total of 20654 statements in the CSL-Daily dataset, the following results are obtained. Figure 11 shows the Bert model training results graph.
\begin{figure}[H]
    \centering
    \includegraphics[width=0.5\linewidth]{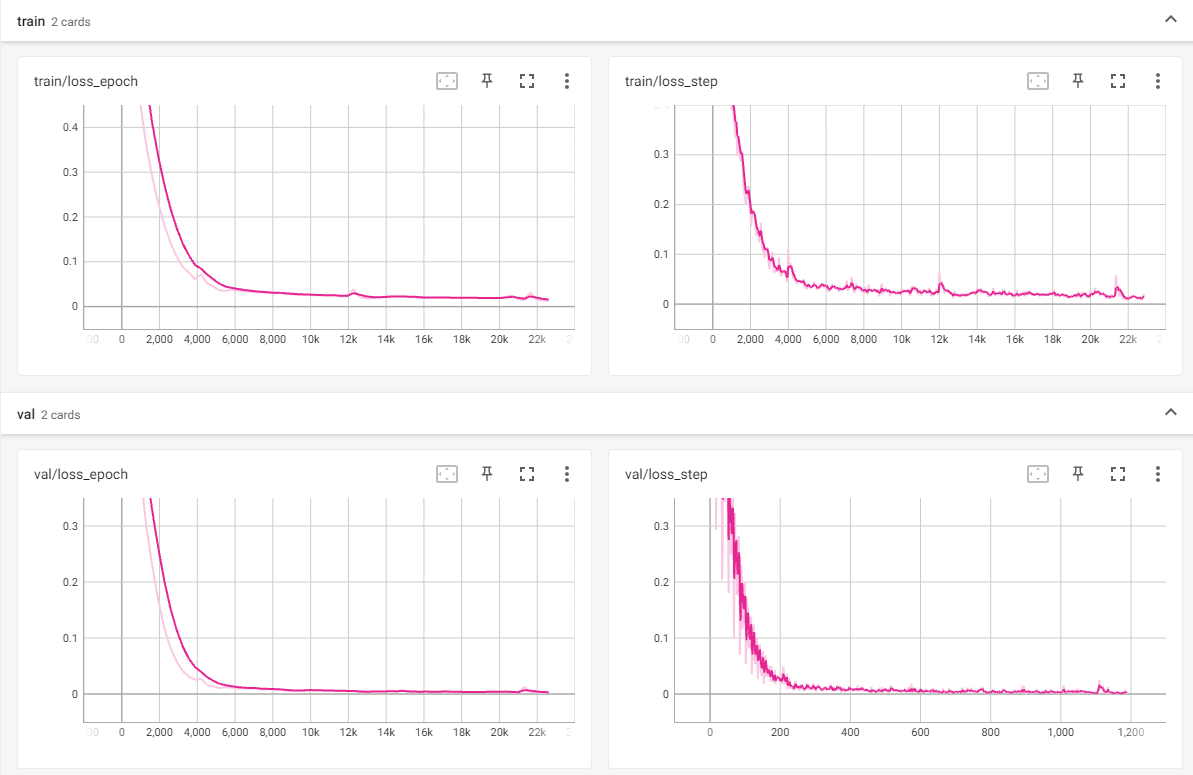}
    \caption{ Bert model training results graph}
    \label{fig:enter-label}
\end{figure}

\section{Disscussion}
Finally, we realize a two-way interactive system for sign language recognition and sign language generation. In sign language recognition, based on the CSL Chinese sign language dataset, the relatively stable and high recognition rate TSM-Resnet50 model is selected, and the continuous sign language sentences are cut and recognized in the sign language video in the test, and the natural order Chinese is generated by adjusting the order.

In the sign language generation, based on the Chinese corpus, word cutting and division of natural utterances are realized by Bert model and adjusted to sign language order, and corresponding videos in the Chinese sign language dataset and generating sign language videos, which realize the two-way recognition and generation of sign language and natural language.

After the test of dividing and recognizing 20 videos of consecutive sign language utterances with less than 10 words, and the test of generating 20 videos of Chinese utterances with less than 20 words in sign language, the accuracy and stability of the two-way process of the system reach a high level.

After future development and improvement, thanks to the convenience and light weight of computer vision, the system will be used in hardware terminals or mobile devices to realize two-way communication between normal groups and hearing-impaired people, and for sign language education to improve the quality of life and happiness of hearing-impaired people, help them better integrate into the general society, and also promote the better development of special education and welfare.

We hope to further improve the accuracy and fault tolerance of recognition by improving the frame cut recognition model and expanding the Chinese sign language corpus in the future, so that it can recognize continuous complex utterances in complex light environments and be put into the welfare business of integrating hearing-impaired people into society as soon as possible.

\vspace{12pt}


\begin{thebibliography}{00}
\bibitem{b1}  A. Vaswani et al., "Attention Is All You Need," arXiv, 2017.

\bibitem{b2} J. Lin et al., "TSM: Temporal Shift Module for Efficient Video Understanding," 2018.

\bibitem{b3} Z. Wang et al., "ACTION-Net: Multipath Excitation for Action Recognition," 2021.

\bibitem{b4} C. Wei et al., "Semantic Boundary Detection With Reinforcement Learning for Continuous Sign Language Recognition," IEEE Transactions on Circuits and Systems for Video Technology, vol. PP, no. 99, pp. 1-1, 2020.

\bibitem{b5} H. Zhou et al., "Improving Sign Language Translation with Monolingual Data by Sign Back-Translation," 2021.

\bibitem{b6} S. Egea et al., "Syntax-aware Transformers for Neural Machine Translation: the Case of Text to Sign Gloss Translation," 2021.

\bibitem{b7} D. Guo et al., "A review of sign language recognition, translation and generation," Computer Science, vol. 048, no. 003, pp. 60-70, 2021.

\bibitem{b8} D. Guo et al., "Connectionist Temporal Modeling of Video and Language: a Joint Model for Translation and Sign Labeling," in Twenty-Eighth International Joint Conference on Artificial Intelligence {IJCAI-19}, 2019.
\end{thebibliography}
\end{document}